\documentclass[conference]{IEEEtran}
\IEEEoverridecommandlockouts
\usepackage{cite}
\usepackage{amsmath,amssymb,amsfonts}
\usepackage{algorithmic}
\usepackage{graphicx}
\usepackage{textcomp}
\usepackage{booktabs}
\usepackage{xcolor}
\usepackage{listings}
\usepackage{float}
\usepackage{tikz}
\usepackage{pgf-pie}
\usepackage{pgfplots}
\usepackage{algorithm}
\def\BibTeX{{\rm B\kern-.05em{\sc i\kern-.025em b}\kern-.08em
    T\kern-.1667em\lower.7ex\hbox{E}\kern-.125emX}}
\begin{document}

\title{LeJOT-AutoML: LLM-Driven Feature Engineering for Job Execution Time Prediction in Databricks Cost Optimization}

\author{\IEEEauthorblockN{Lizhi Ma\IEEEauthorrefmark{2}\IEEEauthorrefmark{3}, Yi-Xiang Hu\IEEEauthorrefmark{2}, Yihui Ren\IEEEauthorrefmark{3}, Feng Wu\IEEEauthorrefmark{1}\IEEEauthorrefmark{2}\thanks{\IEEEauthorrefmark{1}Corresponding author.}, Xiang-Yang Li\IEEEauthorrefmark{2}}
\IEEEauthorblockA{\IEEEauthorrefmark{2}
\textit{University of Science and Technology of China}, Hefei, China\\
\IEEEauthorrefmark{3}
\textit{Lenovo}, Beijing, China\\
\{malizhi,yixianghu\}@mail.ustc.edu.cn, 
renyh10@lenovo.com, 
\{wufeng02,xiangyangli\}@ustc.edu.cn}}
\maketitle

\begin{abstract}
Databricks job orchestration systems (e.g., LeJOT) reduce cloud costs by selecting low-priced compute configurations while meeting latency and dependency constraints. Accurate execution-time prediction under heterogeneous instance types and non-stationary runtime conditions is therefore critical. Existing pipelines rely on static, manually engineered features that under-capture runtime effects (e.g., partition pruning, data skew, and shuffle amplification), and predictive signals are scattered across logs, metadata, and job scripts—lengthening update cycles and increasing engineering overhead. We present LeJOT-AutoML, an agent-driven AutoML framework that embeds large language model agents throughout the ML lifecycle. LeJOT-AutoML combines retrieval-augmented generation over a domain knowledge base with a Model Context Protocol toolchain (log parsers, metadata queries, and a read-only SQL sandbox) to analyze job artifacts, synthesize and validate feature-extraction code via safety gates, and train/select predictors. This design materializes runtime-derived features that are difficult to obtain through static analysis alone. On enterprise Databricks workloads, LeJOT-AutoML generates over 200 features and reduces the feature-engineering and evaluation loop from weeks to 20–30 minutes, while maintaining competitive prediction accuracy. Integrated into the LeJOT pipeline, it enables automated continuous model updates and achieves 19.01\% cost savings in our deployment setting through improved orchestration.
\end{abstract}


\section{Introduction}

Large language models (LLMs)\cite{hurst2024gpt} and agentic reasoning frameworks (e.g., ReAct~\cite{Yao2022ReAct}) have advanced automated code generation and document understanding~\cite{Wang2023LLMAgentsSurvey,Chen2016XGBoost,li2023agent}.
However, enterprise machine-learning (ML) workflows for regression and forecasting still rely on manual feature engineering and brittle glue code, which slows adaptation to workload drift and platform evolution~\cite{Feurer2020AutoSklearn2}.
This challenge is pronounced in Databricks job orchestration, where a small prediction error can translate into repeated mis-provisioning across thousands of daily runs.

LeJOT\cite{LEJOT_2025} is a Databricks-based framework that minimizes execution cost under dependency and latency constraints by selecting resource configurations using an execution-time predictor\cite{Kipf2019MSCN,Park2020Naru}. 
In production, the predictor must generalize across heterogeneous instance types, changing software stacks, and non-stationary data characteristics.
Four obstacles arise in this setting.
First, high-impact performance signals only emerge at runtime~\cite{kim2023dynamic}: scan volume after partition pruning, skew-induced stragglers, shuffle amplification, and executor scheduling effects.
Second, these signals are fragmented across multiple sources (logs, metadata, job scripts, and configuration histories), which complicates the end-to-end feature pipeline.
Third, manual feature engineering demands domain expertise in Spark SQL and platform internals, and the resulting features often lag behind evolving workloads.
Finally, slow retraining and validation cycles yield stale predictors under drift~\cite{Venkataraman2016Ernest}, which degrades orchestration quality and cost efficiency.

To address these limitations, we propose LeJOT-AutoML, an agent-driven ML pipeline that turns the conventional lifecycle\cite{LeDell2020H2OAutoML,Peng2018SLAQ} into a dynamic, self-improving system\cite{he2019automl,feurer2019auto}. 
A Feature Analyzer Agent (FAA) retrieves domain knowledge via retrieval-augmented generation (RAG)~\cite{liu2024rag} and proposes candidate feature templates that map to observable artifacts.
A Feature Extraction Agent (FExA) then invokes a Model Context Protocol (MCP) toolchain~\cite{MCP2024Spec,Schick2023Toolformer} (log parsers, metadata queries, and a read-only SQL sandbox) to materialize both static and runtime-derived features with execution-time validation.
A feedback-enabled Feature Evaluation Agent (FEvA) evaluates feature quality and model performance, and iteratively refines the pipeline to accelerate adaptation under drift~\cite{zhang2023job,wang2021spark,li2023caafe,Venkataraman2016Ernest}.
Two safety gates---a code-completion checker and a data-leakage checker---filter invalid extractors and prevent label leakage.

Our contributions are summarized as follows:
\begin{itemize}
\item \textbf{LLM-powered AutoML pipeline for enterprise job runtime prediction.} We embed LLM agents across analysis, tool invocation, feature extraction, validation, training, and model selection to enable rapid retraining and inference-time feature materialization.
\item \textbf{Agent–tool collaborative feature extraction via MCP.} By combining LLM planning with tool-based execution and verification, LeJOT-AutoML efficiently extracts dynamic features that are inaccessible to purely static analysis.
\item \textbf{Iterative evaluation loop with safety gates.} We introduce a feedback-driven evaluation agent and safety checks (code-completion and data-leakage detection) to improve reliability and drive iterative refinement until predefined criteria\cite{zhang2023job,wang2021spark,li2023caafe} are met.
\end{itemize}

\section{Background and Motivation}
LeJOT~\cite{LEJOT_2025} performs cost-aware orchestration on Databricks by recommending resource configurations that minimize execution cost while satisfying dependency and latency constraints.
The orchestration relies on two coupled components: (i) execution-time estimation under candidate resource allocations and (ii) an optimizer that selects the lowest-cost configuration that meets predicted-time constraints.
Prediction errors therefore propagate directly into orchestration: underestimation violates latency Service Level Objectives (SLOs), while overestimation drives over-provisioning and recurring cost waste.

A central obstacle is that the effective processed data volume of SQL workloads is largely determined at runtime.
Although data volume is a strong predictor of execution time, it does not align with static metadata (e.g., table row counts).
Actual scan and shuffle volumes depend on query logic, data distributions, and optimizer behavior, which invalidates many static proxies.
Static feature engineering thus fails in the following scenarios: 1) Partition pruning: Queries over partitioned tables scan a subset of data; using the total table size overestimates scan volume by orders of magnitude. 2) Join skew: Skewed key distributions concentrate work on a subset of tasks, producing stragglers and violating linear scaling assumptions. 3) Aggregation-induced shuffle: Operators like \texttt{GROUP BY} trigger intermediate shuffles; input table size underestimates network and compute costs under shuffle amplification.

These cases show that ``hidden'' runtime features---selectivity, skew severity, shuffle degree, and stage-level variance---are critical for accuracy yet hard to capture with fixed, hand-written extraction logic.
Moreover, the relevant evidence is distributed: the SQL text encodes logical operators, the execution plan and logs encode physical behavior, and the metadata store encodes schema and partition structure.
A practical pipeline must bridge these sources and update quickly when workloads drift~\cite{Venkataraman2016Ernest}.
LeJOT-AutoML addresses this gap by using LLM agents to synthesize and revise feature extractors grounded in tool-executed evidence (log parsing, metadata queries, and sandboxed SQL analysis), then validating them with safety gates before training.

\section{LeJOT-AutoML Framework}
\subsection{Overview}

\begin{figure*}[htbp]
\centerline{\includegraphics[width=0.95\textwidth]{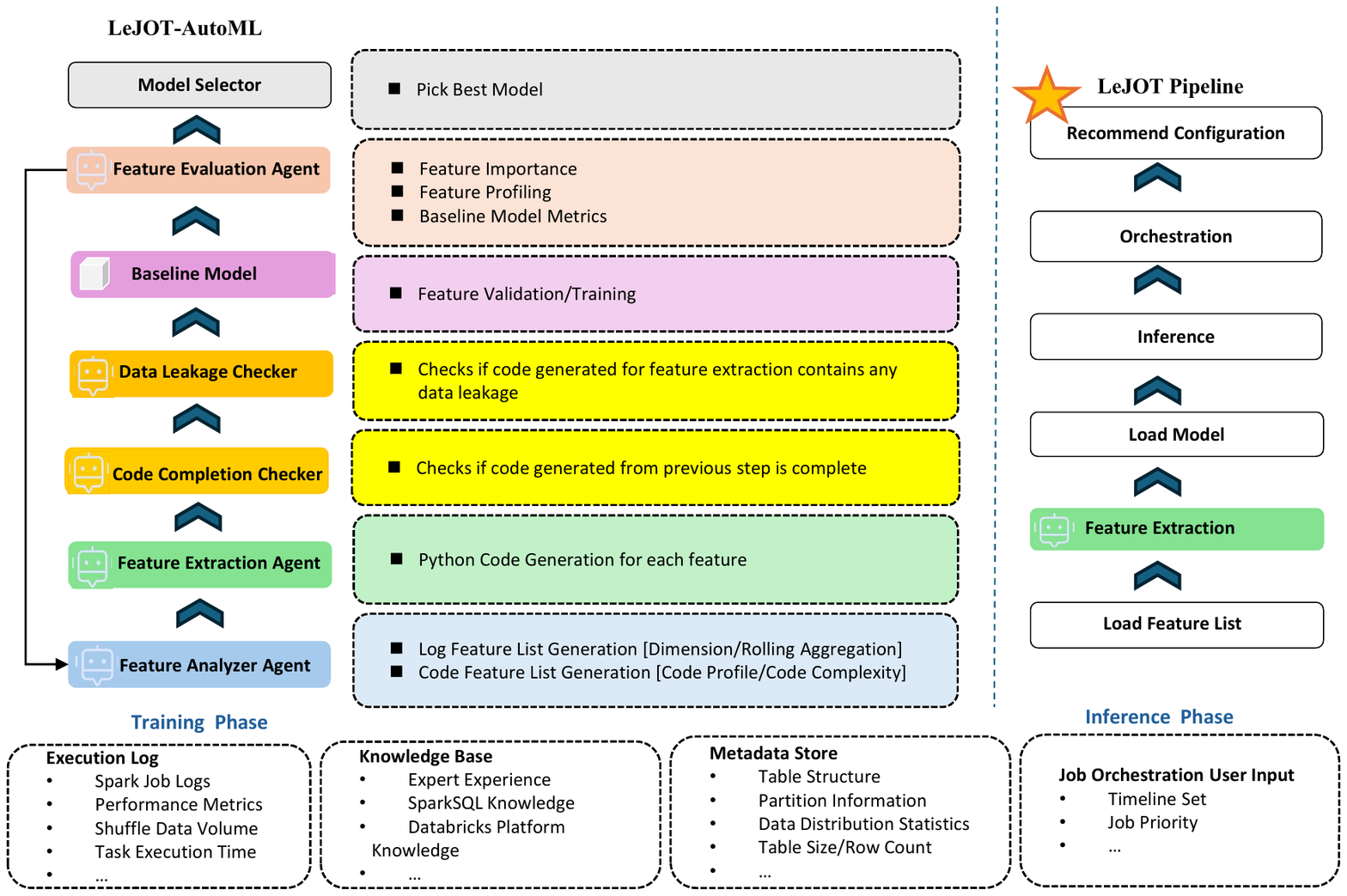}}
\caption{Overview of LeJOT-AutoML. (a) Left: the end-to-end LeJOT-AutoML system. (b) Right: how LeJOT-AutoML integrates into the LeJOT pipeline.}
\label{fig}
\end{figure*}

Figure~\ref{fig} shows the LeJOT-AutoML architecture and its integration into the LeJOT pipeline.
The framework operates in two phases: (i) an automated training phase that generates and validates artifacts,
and (ii) an online inference phase for real-time prediction.

\textbf{Training Phase.}
LeJOT-AutoML forms a closed-loop AutoML pipeline consisting of five core components:
the Feature Analyzer Agent (FAA), Feature Extraction Agent (FExA), a baseline model,
the Feature Evaluation Agent (FEvA), and a model selector.
Two safety gates---a \textit{code-completion checker} and a \textit{data-leakage checker}---ensure that
generated feature-extraction code is executable and does not leak label information.

The training workflow starts with FAA, which ingests three heterogeneous sources:
execution logs (e.g., shuffle volumes and stage/task runtimes), a domain knowledge base (e.g., Spark SQL practices),
and a metadata store (e.g., schema, partitions, and data statistics).
Guided by RAG\cite{lewis2020rag}, FAA proposes a structured feature list.
FExA then generates extraction programs and materializes feature values through an MCP toolchain
(e.g., metadata queries and a read-only SQL sandbox).
After passing safety checks, the resulting feature matrix is used to train baseline and candidate models.
FEvA evaluates feature quality (coverage, skewness), feature utility (importance and redundancy),
and model-level metrics, then emits actionable feedback for refinement.
Finally, the model selector chooses the best-performing model (e.g., XGBoost \cite{Chen2016XGBoost},
LightGBM \cite{Ke2017LightGBM}) and hyperparameters for deployment.

\textbf{Inference Phase.}
For a new job, FAA reuses learned feature templates to determine the required feature set and the corresponding
extraction plan. FExA extracts features in parallel from scripts/metadata and via lightweight sandbox analysis.
The standardized feature vector is fed into the deployed predictor to estimate execution time,
which is then consumed by LeJOT's orchestration algorithm to select a cost-minimizing configuration.
Prediction residuals and feature health signals are logged and fed back into the training loop for continuous updates.

\subsection{System interfaces and design goals}
LeJOT-AutoML sits between raw job artifacts and the downstream orchestration algorithm.
Its inputs consist of (i) static artifacts (job code, configuration, cluster specification, metadata),
(ii) runtime traces (logs and metrics), and (iii) an evolving knowledge base that stores domain rules and ``feature experience''
derived from previous runs. The outputs consist of a versioned feature specification and a deployed predictor
with a reproducible extraction bundle.

We follow four design goals.
\textbf{(1) Low-latency inference:} extraction plans prioritize features with bounded runtime cost, and the toolchain executes
reads under a strict sandbox policy.
\textbf{(2) Safety and governance:} every generated extractor passes syntactic completeness checks and leakage screening before execution.
\textbf{(3) Continuous adaptation:} model retraining is triggered by drift signals or a periodic schedule, reducing staleness in dynamic workloads.
\textbf{(4) Traceability:} each feature records provenance (source, transformation, and collection method), which supports debugging and compliance.

\subsection{Mathematical Formulation}
Let $\mathcal{D}_t = \{d_i\}_{i=1}^{n_t}$ denote the dataset at time $t$,
where each instance $d_i = (d_i^{s}, d_i^{u})$ contains structured and unstructured information,
and let $\mathbf{Y}_t = \{y_i\}_{i=1}^{n_t}$ be the observed execution times.
Given a knowledge base $\mathcal{K}$ and an MCP toolset $\mathcal{T}$,
the agent performs feature analysis $\phi_{\text{analyze}}$ and feature extraction $\phi_{\text{extract}}$.
For each instance, the agent retrieves the domain context via RAG:
\begin{equation}
R_i = \text{RAG}(Q(d_i), \mathcal{K}),
\end{equation}
and determines a job-specific feature set:
\begin{equation}
F_i = \phi_{\text{analyze}}(d_i, R_i).
\end{equation}
Feature values are materialized via tool invocation:
\begin{equation}
\mathbf{x}_i = \{\phi_{\text{extract}}(f, d_i, \mathcal{T}) \mid f \in F_i\}.
\end{equation}
Stacking all instances yields the feature matrix $\mathbf{X}_t = [\mathbf{x}_1, \mathbf{x}_2, \ldots, \mathbf{x}_{n_t}]^{\top}$.
A predictor $M(\cdot;\theta)$ is trained by minimizing a loss $\mathcal{L}$:
\begin{equation}
\theta_{t}^* = \arg\min_{\theta} \mathcal{L}(\mathbf{X}_t, \mathbf{Y}_t; \theta),
\end{equation}
and the deployed model $M_t(\cdot;\theta^*_t)$ predicts $\hat y=M_t(\mathbf{x}_{\text{new}})$ for a new job $d_{\text{new}}$.

\textbf{Cost-aware feature selection.}
Online extraction introduces a latency budget that constrains the usable feature set.
Let $c(f, d)$ denote the extraction cost of feature $f$ on job artifact $d$, and let $B$ be a per-request budget.
FAA therefore targets feature sets that jointly improve prediction accuracy and satisfy runtime constraints:
\begin{equation}
\begin{aligned}
    F_{\text{new}} = \arg\min_{F \subseteq \mathcal{F}} \ \mathbb{E}\!\left[\ell\!\left(M(\mathbf{x}_F), y\right)\right]
+ \lambda \sum_{f \in F} c(f, d_{\text{new}})\quad\\
\text{s.t. } \sum_{f \in F} c(f, d_{\text{new}}) \le B,
\end{aligned}
\end{equation}
where $\mathcal{F}$ is the global feature universe, $\mathbf{x}_F$ denotes the subvector restricted to $F$,
and $\lambda$ controls the accuracy--latency tradeoff.

\textbf{Safety constraints.}
Each generated extractor program $p_f$ is executed only if it passes two gates:
a syntactic completeness predicate $g_{\text{cc}}(p_f)=1$ and a leakage predicate $g_{\text{dl}}(p_f)=1$.
These gates impose hard constraints on feasible feature sets:
\begin{equation}
\forall f \in F:\quad g_{\text{cc}}(p_f)=1 \ \wedge\  g_{\text{dl}}(p_f)=1.
\end{equation}

\textbf{Continuous updates.}
When new data $\mathcal{D}_{\text{new}}$ arrives, the system updates
$\mathcal{D}_{t+1} = \mathcal{D}_t \cup \mathcal{D}_{\text{new}}$,
$\mathbf{Y}_{t+1} = \mathbf{Y}_t \cup \mathbf{Y}_{\text{new}}$, and retrains:
\begin{equation}
\theta_{t+1}^* = \arg\min_{\theta} \mathcal{L}(\mathbf{X}_{t+1}, \mathbf{Y}_{t+1}; \theta).
\end{equation}
This cycle is triggered periodically (e.g., daily) or by drift signals to maintain reliable predictions under workload evolution.

\subsection{Implementation Details}
We summarize implementation choices that make LeJOT-AutoML practical in an enterprise setting, focusing on (i) the MCP tool interface, (ii) execution and caching policies, and (iii) feedback and drift handling.

\paragraph{MCP tool interface and outputs} Each MCP tool exposes a restricted, typed interface and returns structured outputs (e.g., JSON-like records) that can be deterministically transformed into features. We group tools into three categories: (i) metadata tools (schema, partition layout, table statistics, cluster configuration), (ii) log/trace tools (stage/task timing, shuffle read/write, spill metrics, failure reasons), and (iii) sandbox tools that execute read-only SQL queries or lightweight plan inspection under strict policies. For robustness, each extractor emits a typed feature schema (name, type, default value, provenance), and the Feature Extraction Agent validates tool outputs against the schema before materialization.

\paragraph{Execution policy (safety, determinism, parallelism)} All generated extractors run in a sandboxed environment with an allowlist of libraries and tool calls. The code-completion checker blocks extractors with missing imports, undefined variables, or unresolved tool outputs. The data-leakage checker enforces availability: a feature must be computable from information available before a scheduling decision (e.g., job scripts and historical traces), and cannot rely on post-run artifacts. For efficiency, independent tool calls are executed in parallel when dependencies permit. We also bound inference latency by capping per-job sandbox queries and enforcing timeouts.

\paragraph{Caching and versioning} To avoid repeated reads for recurring jobs, we cache intermediate tool outputs and materialized feature values. Cache keys are tuples of (job signature, data snapshot identifier, feature version, tool version), enabling safe reuse and incremental retraining by re-materializing only affected features. Each deployed model is packaged with a versioned feature specification and extractor bundle to ensure online inference replays training-time transformations. 

\paragraph{Feedback signals and drift-triggered updates} LeJOT-AutoML logs prediction residuals, feature health signals (missingness, outliers, schema mismatches), and extraction latency per tool modality. These signals drive periodic refresh (e.g., daily retraining) and drift-triggered refresh when residuals or feature distributions exceed thresholds. During retraining, FEvA summarizes failure modes (unstable features, high-cardinality categoricals, redundancy) and feeds concise guidance to FAA/FExA to refine feature specifications and extraction plans.

\section{Design of Core Module Functions}

\subsection{Feature Analyzer Agent (FAA)}
The FAA determines the candidate feature space and therefore the accuracy ceiling of the predictor.
It takes as input: (i) the task objective (execution-time prediction for cost-aware orchestration),
(ii) supplementary artifacts (job scripts, configuration, historical logs, cluster and table metadata),
(iii) constraints (collection cost, privacy policy, and access scope), and (iv) an output schema.
Using this information, FAA performs two functions.

\textbf{(1) Context recovery via RAG.}
FAA formulates queries over $\mathcal{K}$ to retrieve Spark SQL and platform knowledge that clarifies which
runtime behaviors dominate performance. Retrieved context is then grounded against job artifacts to avoid generic suggestions.

\textbf{(2) Feature specification synthesis.}
FAA outputs a list of \emph{feature specifications} rather than raw names. Each specification records:
\emph{name}, \emph{type} (numerical/categorical/text-derived), \emph{source} (log/metadata/code),
\emph{extraction plan} (tool calls and transformations), and \emph{expected cost} and \emph{refresh frequency}.
This schema supports traceability, caching, and downstream parallel extraction.

\subsection{Feature Extraction Agent (FExA)}
The FExA materializes the proposed features via an agent--tool collaborative architecture.
The LLM performs planning and program synthesis, while the MCP toolchain executes and verifies retrieval steps
through a constrained interface \cite{MCP2024Spec}.
FExA follows a four-stage extraction pipeline:

\begin{itemize}
    \item \textbf{Static extraction:} parses job scripts and metadata to obtain invariant features
    (e.g., operator counts, join patterns, table statistics, partition layouts).
    \item \textbf{Runtime materialization:} invokes log parsers and the read-only SQL sandbox to collect
    runtime-derived signals (e.g., stage imbalance, shuffle amplification, pruning effectiveness).
    \item \textbf{Normalization and encoding:} maps heterogeneous outputs into a unified feature vector through
    scaling, one-hot encoding, and text vectorization when needed.
    \item \textbf{Data-quality checks:} flags missing values, outliers, and schema mismatches, then emits
    repair actions (fallback features, default values, or re-execution with tightened queries).
\end{itemize}

To meet inference latency goals, FExA executes independent extractors in parallel and uses caching keyed by
(job signature, feature version, and data snapshot), which reduces repeated reads across similar recurring jobs.

\subsection{Feature Evaluation Agent (FEvA)}
The FEvA performs a multi-level assessment to decide which features enter the deployed model.
It aggregates three families of signals:

\textbf{Feature health.}
For each feature, FEvA measures coverage (missing rate), stability (variance under similar jobs),
and distribution shifts across time windows, identifying brittle or non-stationary features.

\textbf{Feature utility.}
FEvA estimates importance and redundancy using model-based attribution (gain/SHAP-style summaries from tree models)
and correlation screening, which reduces collinearity and overfitting risks.

\textbf{End-to-end impact.}
FEvA evaluates the marginal impact of candidate features through ablations on baseline models
and reports deltas in MAE/RMSE and $R^2$. The output is a structured feedback packet that instructs FAA/FExA
to refine extraction plans, drop unstable features, or propose additional domain-grounded interactions.

\begin{table}[t]
\centering
\caption{Feature Diversity Comparison}
\label{tab:feature_comparison}
\begin{small}
\resizebox{\columnwidth}{!}{
\begin{tabular}{lcc}
\hline
\textbf{ } & \textbf{AutoML}  & \textbf{Manual}  \\
\hline
Number of Features & 200+  & 40+ \\
\hline
\multicolumn{1}{l}{Feature Types} &
\begin{tabular}[c]{@{}l@{}}- Log Profiling Features \\ - Historical Time Series Data \\ - Driver Node Historical Data \end{tabular} &
\begin{tabular}[c]{@{}l@{}}- Node Configuration Historical Data  \\ - Log Profiling Features \end{tabular}  \\
\hline
\end{tabular}
}
\end{small}
\end{table}

\begin{table}[tbp]
\centering
\caption{Top-5 Feature Importance for AutoML vs. Manual}
\label{tab:feature_importance_bar_simple}
\resizebox{\columnwidth}{!}{\begin{tabular}{lcc}
\hline
\textbf{Feature Name} & \textbf{AutoML (\%)} & \textbf{Manual (\%)} \\
\hline
duration\_seconds\_lag\_1           & 28.8 & --   \\
vcpu\_lag\_1                        & 24.6 & --   \\
duration\_seconds\_shifted\_avg\_last\_3\_runs & 8.0 & -- \\
DBU\_lag\_1                         & 7.8  & --   \\
Memory\_lag\_1                      & 3.6  & --   \\
vcpu\_ratio                         & --   & 16.2 \\
vmemory\_ratio                      & --   & 13.4 \\
total\_memory\_changed              & --   & 8.0  \\
total\_cpu\_changed                 & --   & 7.9  \\
worker\_flexibility\_ratio          & --   & 4.1  \\
\hline
\end{tabular}}
\end{table}

\subsection{Safety gates and model selection}
LeJOT-AutoML executes generated code under strict safety gates.
The \textbf{code-completion checker} verifies that each extractor program is syntactically complete,
imports only approved libraries, and returns values conforming to the feature schema.
The \textbf{data-leakage checker} enforces temporal and semantic isolation between features and labels,
rejecting extractors that directly access the target (execution time) or indirectly derive it from post-run artifacts.

After FEvA produces evaluation summaries, the \textbf{model selector} searches a bounded candidate set of algorithms
and hyperparameters, trains candidates on the versioned feature matrix, and selects the final configuration for deployment.
The selected model is packaged together with its feature specification and extractor bundle, ensuring that online inference
replays the same transformations used during training.

\section{Experiments}

% \subsection{Computing Environment}
We evaluate LeJOT-AutoML on enterprise Databricks workloads by comparing \textit{AutoML} (LLM+MCP automated feature engineering) with \textit{Manual} feature engineering. The pipeline follows an analysis–act–validate loop: the LLM parses unstructured job artifacts (scripts and logs), the MCP toolchain materializes runtime-derived signals, and safety checks validate extracted features. We use 5-fold cross-validation to estimate generalization and conduct ablation studies to quantify the contributions of different feature sources. We report feature diversity, prediction metrics (MAE, MAPE, $R^2$), per-module runtime, and end-to-end cost savings in LeJOT.

Experiments were conducted on a single machine with an Intel Core Ultra 7 165U CPU, 32 GB RAM, and Windows 11 Professional. We used Qwen-235B\cite{yang2025qwen3} for agent reasoning and feature/code synthesis, and trained the execution-time predictor using XGBoost\cite{Chen2016XGBoost}.

\begin{table}[tbp]
\centering
\caption{Comparison of AutoML and Manual Feature Extraction Approaches}
\label{tab:compare}
{
\begin{tabular}{lcc}
\hline
\textbf{Metrics} & \textbf{AutoML} & \textbf{Manual}  \\
\hline
R$^2$ & 0.81& 0.91 \\
MAPE & 20.13$\%$ & 19.49$\%$  \\
MAE & 123.29 & 78.94\\
Time  & 20$-$30 min (3 iterations) & 1 Month  \\
\hline
\end{tabular}
}
\end{table}

\begin{table}[tbp]
\centering
\caption{Inference results for a representative job under two feature engineering methods}
\label{tab:inference result}
{
\begin{tabular}{lrr}
\hline
\textbf{Compute Machine} & \textbf{AutoML (s)} & \textbf{Manual (s)}  \\
\hline
Standard\_F4s & 167 & 206 \\
Standard\_F16s & 154 & 82 \\
Standard\_E16\_v4  & 162 & 78\\
Standard\_F16\_v4\_Photon & 147 & 52  \\
\hline
\end{tabular}
}
\end{table}

\subsection{Manual vs. AutoML}

AutoML synthesizes more than 200 features spanning log profiling, time-series statistics, and driver-node history, whereas manual engineering yields around 40 features, largely derived from node-configuration history (Table~\ref{tab:feature_comparison}). The two pipelines also surface markedly different top-ranked features (Table~\ref{tab:feature_importance_bar_simple}), indicating that AutoML emphasizes temporal and workload-dependent signals beyond the static resource ratios that dominate many hand-crafted designs.

A key source of this difference lies in the availability of control-plane context. Manual features leverage historical cluster sizing decisions, instance family transitions, and configuration--price mappings. In our current deployment, the MCP toolset provides configuration snapshots but does not expose configuration-change trajectories or pricing context with comparable fidelity. Consequently, FAA focuses on runtime-derived evidence grounded in logs and plan inspection, and it under-represents certain resource-history signals that manual engineering explicitly encodes. Extending the tool interface with configuration-change logs and instance-specification/price knowledge is a clear path to improving resource awareness and narrowing this gap.

\begin{table}[tbp]
\centering
\caption{Execution time (in seconds) for each agent node across five experimental runs.}
\label{tab:node_times}
\resizebox{\columnwidth}{!}{
\begin{tabular}{lrrrrr}
\hline
\textbf{Agent} & \textbf{Run 1} & \textbf{Run 2} & \textbf{Run 3} & \textbf{Run 4} & \textbf{Run 5} \\
\hline
Feature Analyzer & 260.60& 233.27& 263.29& 223.47& 298.12\\
Feature Extractor & 107.80& 92.71& 105.25& 115.79& 137.87\\
Code Completion Checker & $\textless$0.01& $\textless$0.01& $\textless$0.01& $\textless$0.01& $\textless$0.01\\
Data Leakage Checker & 199.89& 172.21& 209.45& 210.34& 325.60\\
Evaluation Agent & 33.39& 38.85& 43.62& 49.66& 38.80\\
Model Selector & 27.12& 20.87& 34.56& 24.93& 32.43\\
\hline
\end{tabular}
}
\end{table}

From an engineering-effort perspective, AutoML completes three end-to-end iterations within 20--30 minutes, whereas manual feature design typically requires about one month (Table~\ref{tab:compare}). Manual features achieve higher predictive accuracy ($R^2$=0.91 vs.\ 0.81), but AutoML delivers competitive performance at a small fraction of the development cost. A representative case study (Table~\ref{tab:inference result}) further highlights the trade-off: when instance types are upgraded and Photon is enabled, the manual model's predictions shift substantially, while the AutoML model exhibits a smaller response. This pattern suggests that AutoML currently does not fully capture the direct effect of resource upgrades, consistent with its limited access to configuration-history and pricing signals.

\subsection{Additional results}
Module-level runtime (Table~\ref{tab:node_times}) shows that FAA and the data-leakage checker dominate end-to-end latency, reflecting the cost of LLM reasoning and semantic verification. Over three FEvA iterations, MAE decreases from 247.95 to 145.64 and $R^2$
 improves from 0.61 to 0.81 (Table~\ref{tab:evaluation2}), validating the effectiveness of the feedback loop. Integrated into LeJOT, AutoML achieves 19.01\% cost savings (Table~\ref{tab:cost_saving}), demonstrating practical value even with a modest accuracy gap.

\begin{table}[tbp]
\centering
\caption{Metrics of the baseline XGBoost model over three iterations of evaluation}
\label{tab:evaluation2}
\footnotesize
\resizebox{\columnwidth}{!}{
\begin{tabular}{lccccc}
\hline
\textbf{Metric} & \textbf{First Iteration } & \textbf{ Second Iteration } & \textbf{Third Iteration }  \\
\hline
MAE & 247.95 & 172.09 & 145.64  \\
MAPE (\%) & 36.28 & 25.18 & 21.31   \\
R$^2$ & 0.61 & 0.80 & 0.81 \\
\hline
\end{tabular}
}
\end{table}

\begin{table}[tbp]
\centering
\caption{Cost Saving Rate Comparison Between Solutions}
\label{tab:cost_saving}
\footnotesize
\resizebox{\columnwidth}{!}{
\begin{tabular}{lcccc}
\hline
\textbf{Solution} & \textbf{Throughput (k/s)} & \textbf{Initial Cost (\$k)} & \textbf{Final Cost (\$k)} & \textbf{Cost Saving Rate} \\
\hline
AutoML    & 0.08 & 52.6 & 42.6 & 19.01\% \\
Manual ML  & 0.12 & 52.6 & 37.9 & 27.94\% \\
\hline
\end{tabular}
}
\end{table}

\section{Conclusion and Discussion}
We presented LeJOT-AutoML, an LLM-driven framework for automated feature engineering in Databricks job execution-time prediction. By integrating LLM agents with an MCP toolchain, the system expands the feature space to include hard-to-observe runtime signals and compresses the feature-engineering cycle from months to minutes. Although manual feature engineering still delivers better generalization across hardware configurations ($R^2$=0.91 vs. 0.81), LeJOT-AutoML provides a scalable, low-maintenance alternative that enables continuous learning and achieves 19.01\% cost savings in LeJOT. Future work will focus on improving resource awareness by incorporating richer configuration- and runtime-level indicators of execution and data-movement behavior.

\section*{Acknowledgments}

The research is partially supported by Innovation Program for Quantum Science and Technology 2021ZD0302900 and China National Natural Science Foundation with Nos. 62132018 and 62231015, ``Pioneer'' and ``Leading Goose'' R\&D Program of Zhejiang, 2023C01029, and 2023C01143, Anhui Provincial Natural Science Foundation under Grant 2208085MF172, and the USTC Kunpeng-Ascend Scientific and Educational Innovation Excellence Center.

\bibliographystyle{IEEEtran}
\bibliography{refer}
\end{document}